# A Novel Two-Staged Decision Support based Threat Evaluation and Weapon Assignment Algorithm

Asset-based Dynamic Weapon Scheduling using Artificial Intelligence Techinques

Huma Naeem, Asif Masood, Mukhtar Hussain, Shoab A. Khan
Department of Computer Science, National University of Science and Technology, Rawalpindi, Pakistan
jinnny@gmail.com , amasood@mcs.edu.pk, mukhtar.dr@gmail.com, kshoab@yahoo.com

*Abstract—. Surveillance control and reporting (SCR) system for air threats play an important role in the defense of a country. SCR system corresponds to air and ground situation management/processing along with information fusion, communication, coordination, simulation and other critical defense oriented tasks. Threat Evaluation and Weapon Assignment (TEWA) sits at the core of SCR system. In such a system, maximal or near maximal utilization of constrained resources is of extreme importance. Manual TEWA systems cannot provide optimality because of different limitations e.g. surface to air missile (SAM) can fire from a distance of 5Km, but manual TEWA systems are constrained by human vision range and other constraints. Current TEWA systems usually work on target-by-target basis using some type of greedy algorithm thus affecting the optimality of the solution and failing in multi-target scenario. his paper relates to a novel two-staged flexible dynamic decision support based optimal threat evaluation and weapon assignment algorithm for multi-target air-borne threats.*

*Keywords- Optimization Algorithm; Threat Evaluation (TE) and Weapon Assignment (WA) Algorithm (TEWA); Decision Support System (DSS); Stable Marriage Algorithm (SMA); Cybernetics Application, preferential and subtractive defense strategies.*

## I. INTRODUCTION

Defense of a country is of supreme importance in this technology saturated age. Automated and semi-automated defense oriented tools can be seen as a military response to this technology pressure. Digitized surveillance control and reporting (SCR) system for air threats is one such system. SCR system corresponds to air and ground situation management/processing along with information fusion, communication, coordination, simulation and other critical defense oriented tasks. Threat Evaluation and Weapon Assignment (TEWA) sits at the core of SCR system. Figure 1 shows graphical representation of integrated digitized surveillance process.

TEWA is a complex system that maintains an ongoing interaction with a non-deterministic and dynamic environment. It includes continuous intelligent threat detection, identification and evaluation coupled with resource allocation in real time and in response to a variety of events taking place in the environment. The situation is more complex when there are multiple potential threats. TEWA can be divided into two sub-processes i.e. threat Evaluation (TW) and Weapon Assignment (WA). A closed loop TEWA process is not only NP-Complete but also, discrete, dynamic, non-linear, stochastic, and large scale in terms of increased number of WS and targets [3], it is difficult to solve it optimally when number of threats and weapons is large, as computation time of the solution increases rapidly with the size of problem. For an efficient TEWA system there is a need to create a balance between usefulness and effectiveness of weapon systems (WSs) [2], [3].

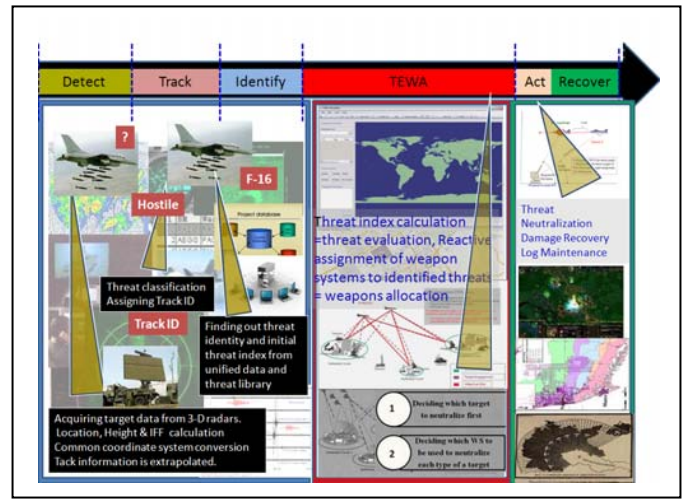

Figure 1: Integrated Defense System

Manual TEWA systems cannot provide optimality because of little amount of information available and established situational picture (limited awareness and information), operator's limitations like vision range constraint, experience, observation, understanding of the situation and mental condition. This is a known fact that humans are prone to errors especially in stressful conditions. As a consequence in military domain, when facing a real attack scenario, an operator may fail to come up with an optimal strategy to neutralize targets. This may cause a lot of ammunition loss with an increased probability of expensive asset damage. Moreover, most semi-automated TEWA systems usually work on target-by-target basis using some type of greedy algorithm thus affecting the optimality of the solution and failing in multi-target scenario [4]. Figure 2 shows graphical representation of TEWA along with two critical concerns found in multi-target scenarios.

This paper relates to the design, simulation and analysis of a novel two-staged flexible dynamic decision support based optimal threat evaluation and weapon assignment algorithm for multi-target air-borne threats. The algorithm provides a near optimal solution to the defense resource allocation problem while maintaining constraint satisfaction. The model used is kept flexible to compute the most optimal value for all classes of parameters.

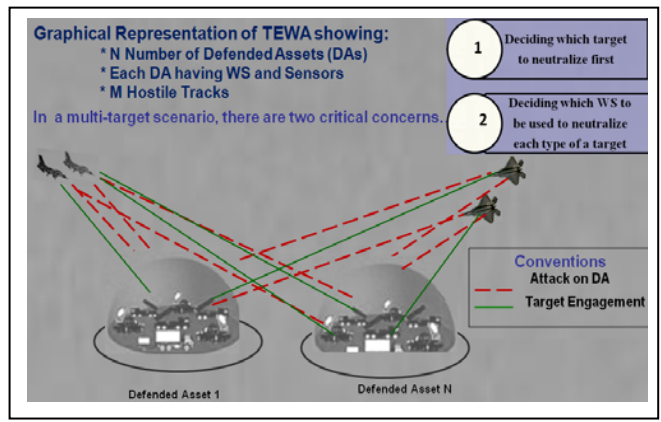

Figure 2: Graphical representation of TEWA along with two critical concerns found in multi-target scenarios

Proposed solution tends to improve the optimality of the solution and assigns the most suitable values to critical parameters sued in TEWA decision making at the expense of extra computation. Section 2 of this paper presents literature survey. Section 3 covers our approach to TEWA along with parameter classification. The performance of proposed model is compared with an alternative greedy algorithm and the strengths and weaknesses of proposed algorithm are briefly discussed in section 4. Section 5 finally concludes this report.

## II. LITERATURE SURVEY

TEWA is a complex system, having maximal or near maximal utilization of constrained resources as one of extreme important external factor. Main purpose of TEWA is to address two critical concerns related to multi-target scenarios i.e. deciding the order in which threats should be neutralized and deciding which WS to be used to neutralize a particular target. First concern relates to TE while weapon selection decision is the task of WA process. The subsequent part of this section discusses these two processes in detail. Figure 3 shows general classification of TE and WA models along with proposed hybrid solution and an outline of pre-processing phase.

Tin G. and P. Cutler classified parameters related to TEWA into critical and sorting parameters, in [4]. We have amended this classification a little and divided the main parameters used in TEWA processing into three overlapping categories i.e.

- **Triggering Parameters:** This class of parameters represents different thresholds used in TEWA system, to initiate a certain function. For example initial threat index is used to trigger TEWA process, status of a Defended Asset (DA) and Weapon System (WS) to be allotted (like Free to Fire, On Hold and Tight).
- **Sorting Parameters:** Parameters belonging to this category are used to rank the threats, DAs and WSs.

Main parameters include threat priority – that sorts the targets from the most threatening to the least threatening, Opportunity Index - a value indicating intent and capability of a target to inflict injury to a DA, calculated using Intent parameters and Capability parameters mentioned in section 2, kill capability (K.C) of a DA (a probability assigned to each DA based on the capability of a DA to destroy a particular type of threat) , lethality Index, condition (Up, Down, Destroyed) and status of a WS, time to DA, time to WS.

- **Scheduling Parameters:** This class refers to the parameters that are used to assign a target either to a DA or to a WS. For example, weight of DA-Target pair (parametric equation using K.C in combination with Time to DA and load on DA), Weight of WS-Target pair (Time to WS in combination with Required elevation, Maximum elevation of a WS, Lethality Index, Stabilization time and Rate of Fire (ROF) of a WS

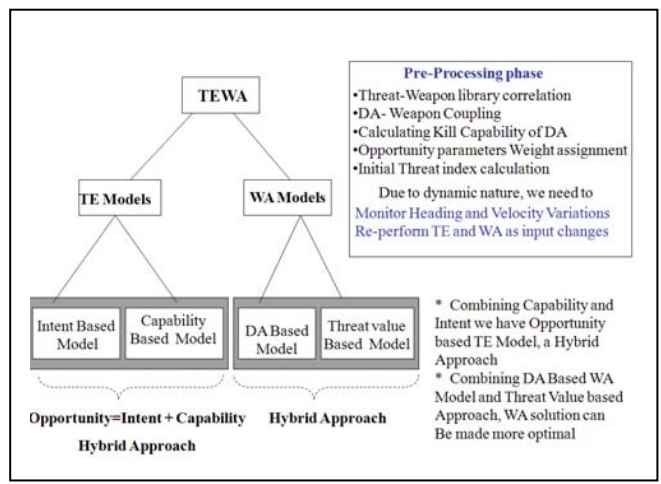

Figure 3: Hybrid two-stage proposed solution along with an outline of pre-processing phase

### A. Threat Evaluation (TE)

TE relates to threat ranking from the most threatening to the least threatening to decide the order in which threats should be neutralized. It consists of multiple systematic and operational activities designed to identify, evaluate, and manage anything that might pose a threat to identifiable targets [5]. TE is highly dependent on established situational picture i.e. target's current state estimates along with available contextual information (location of the defended assets (DAs) and attributes of Weapon Systems (WSs)) [6]. It can be seen as an ongoing two step process, where in first step, we determine if an entity intends to inflict damage to the defending forces and its interests, and in second stage we rank targets according to the level of threat they pose [7]. Typically threat evaluation has been seen from two perspectives:

- Capability based TE Model – these models rank targets according to their capability index, where capability index defines the ability of the target to inflict damage to a DA.

- Intent based TE Model – these models estimate intent of a target to calculate threat index. Intent refers to the will of a target to inflict damage to a DA.

According to the literature, different parameters can be used to estimate the capability and intent of an identified threat for example target type, speed, direction, weapon type and envelope etcetera can be used to estimate capability of a threat while, heading (bearing and course), velocity, altitude and speed etcetera, can help predict intent of an identified threat [7]. Combining these two we have opportunity based threat evaluation.

### B. Weapon Allocation (WA)

Primitive battlefield modeling can be done using basic mathematical functions and rules. In reality, there can be N number of Defended Assets (DAs) having M number of Weapon systems (WSs) of different types. Each WS usually has its own lethality index, priority, rate of fire, field of fire, elevation angle and other parameters. Not all WSs have ability to neutralize every kind of threat. So, while selecting best WS to encounter a threat, there is a need to consider capability and suitability of each WS as applicable to each scenario. DAs too have important parameters associated with them for example each DA may have its own priority and vulnerability index. Based on these values different parametric values and WSs are assigned to each DA. WA corresponds to selecting the best available weapon system to neutralize a particular target i.e. selecting which WSs to be used to neutralize a particular target in a multi-threat scenario. [8] – [14] provide a good literature on WA problem. TEWA, being a real time system is subject to uncertainties and hard set of external constraints. Due to dynamic ever-changing environment, assignment problems might need to be re-solved, this makes scheduling problem even more complex to design and implement [7].

Design of WA algorithm depends on defensive strategy to be implemented i.e. the defense may either want to maximize the total expected value of the DAs, or the defense may want to minimize the total expected surviving value of the targets that survive all weapon engagements. The first approach using max function is known as preferential defense strategy while the later one is known as subtractive defense strategy [15]. According to Hosein [15], subtractive defense strategy can be seen as a special case of preferential defense strategy when numbers of threats per DA are small. If we increase the number of threats directed to each DA, subtractive defense fails. On the contrary preferential strategy is highly sensitive to uncertainty level of input data [15].

Execution models for WA can fall into two categories, the ones that take into account notion of time i.e. dynamic execution model and the models without the notion of time i.e. static models that include processes defined over a single time horizon [16]. Using dynamic approach we can make multiple engagements in stages by observing the outcomes of previous engagement before making any further engagements [17].

From implementation point of view each of this stage can iterate through Boyd's Observe-Orient-Decide-Act (OODA) loop [6]. Figure4 shows OODA loop for WA.

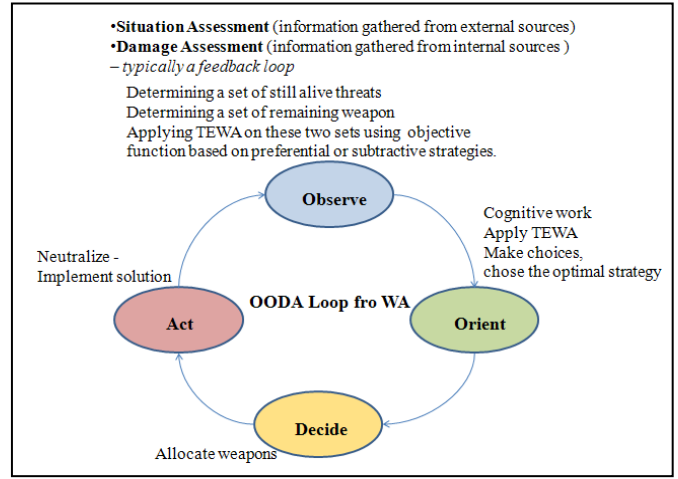

Figure 4: Boyd's OODA loop for WA

### III. OUR APPROACH

In this section, a new two staged model of Threat Evaluation and Weapon Assignment is proposed. The assignment problem is formulated as an optimization problem with constraints and then solved by a variant of many-to-many Stable Marriage Algorithm (SMA).

### A. PreProcessing Phase

We present a Decision Support based system, consisting of correlated threat and weapon libraries. To formulate the TEWA problem, we need the information about the kill probabilities and lethality index for each weapon-target pair. This information comes from threat and weapon libraries. Our solution is based on two way correlation between weapon and threat library. Kill probabilities and lethality index are dependent not only on threats but also on WSs and load on WSs. Threat library holds data for all possible type of threats like Ground-attack aircrafts, Fighter aircrafts, Helicopters, Interceptors, Reconnaissance, Trainer and Transport aircrafts etcetera. Weapon library on the other hand, holds data for different type of weapons in hand like Cannons, Rockets, Ground Missiles, Smart Bombs, Free Fall Bombs, and Low Level Attack Bombs etcetera. To make TE process optimal we use weapon effectiveness based threat/weapon correlation database, that defines the highest priority, most effective, weapon to counter expected threats, thus ensuring that the selected weapons have the capability to neutralize particular target, improving the cost and value of the assignment. TE and WA processes use different parameters. All these parameters have their own significance and thus can be used to make different weighted parametric equations in TE and WA process. To get optimal weight value for each parameter; weights can be kept configurable to see their impact in different offline scenarios. To assign values to these parameters fuzzy logic is used restricting values of parameters between 0 and 1. The solution considers one or more type of WSs possessed by a set of DAs against a set of threats k, and allows for multiple target assignment per DA. The model is kept flexile to handle unknown type of threats.

## B. Threat Evaluation Model

Suppose the defense has i numbers of DAs to be guarded against potential threats. Where each DA has its own set of one or more type of WSs; suppose the total number of WSs is j. Assume DAs can be attacked by k number of threats of any kind and capability in any order. The main objective of this process is to rank the targets according to their opportunity index and assign each target to the most suitable DA.

We model DAs with circles and threats with line segments. Using circle line intercept equations, we compute the points of intersection between each threats and DA. Let the DA be shown by equation 1.

$$(x - x_0)^2 + (y - y_0)^2 = r^2 \quad (1)$$

Where x0, y0 are center points of DA and r shows the radius of circle representing $DA_i$.

Extending threat velocity vector we get a line represented by equation 2.

$$y = mx + c \quad (2)$$

To assign threats to a DA, we need to calculate proximity parameters, intent parameters and capability parameters of threats and DAs. Threat ordering is important because we want to neutralize the most threatening targets as early as possible. After assigning initial threat index, TE processes all identified threats to calculate refined threat index based on opportunity (intent and capability) and proximity parameters. This refined threat index specifies the order in which threats should be processed for WA. Proposed solution uses Defended Asset (DA) based TE by applying (one-to-one) stable marriage algorithm (SMA) with weighted proposals between threats and DAs. For each threat, list of matching capable DAs is searched on the basis of kill capabilities (index), status of DA (Free to Fire, On Hold, and Freeze) and priorities of DAs. So, a threat k, proposes to only those DAs that have capability high enough to neutralize it. Each proposed DA, looks for proposal acceptance feasibility and responds accordingly. Each DA has its own vulnerability index, priority and other parameters. Weights are assigned to each DA on the basis of DA priority and threat related parameters like Time to DA (TDA), heading, velocity etc. For each threat k, the proposal with the maximum weight is processed first. If situation allows $DA_i$ to accept this proposal, threat is assigned to proposed DA, else the next proposal highest weighted proposal is sent. Mostly TDA calculations are based on constant speed that is quite unrealistic. Our proposed algorithm caters for this deficiency by calculating speed and velocity along different axis at different time stamps, making it scale up its efficiency for scenarios modeling track maneuvering. Once a threat is assigned to DA, WA process can start.

Proximity parameters are closely related to intent parameters. So we calculate earliest DA point of intersection (POI) for each threat and DA that has Kill probability high enough to neutralize assigned threat. Kill probability K.P of a $DA_i$ is shown by (3) where i = {1,2,3….. $N_{DA}$}; $N_{DA}$= Number of DAs.

$$\left(\prod_{k=1}^{K}\left(1 - \left((W_I * II_k + W_{cI} * cI_k + W_L * Load_j) * C_{j,k}\right)^{B_{i,j}}\right)\right) \quad (3)$$

Where

$K \stackrel{def}{=}$ Number of Threats

$W_I \stackrel{def}{=}$ Weight Assigned to Intent Parameters

$II_k \stackrel{def}{=}$ Intent index of threat k

$W_{CI} \stackrel{def}{=}$ Weight Assigned to Capability Parameters

$CI_k \stackrel{def}{=}$ Capability index of threat k

$W_L \stackrel{def}{=}$ Weight Assigned to Load Parameter

$Load_k \stackrel{def}{=}$ Load on $WS_j$; j ∈ {1,2,... $N_{WS}$}; $N_{WS}$= Number of WSs ∈ $DA_i$.

To calculate POIs we expand equation 1 as:

$$x^2 + x_o^2 - 2xx_0 + y^2 + y_o^2 - 2yy_0 = r^2 \quad (4)$$

Putting (2) in (4) we have,

$$x^2 + x_o^2 - 2xx_0 + (mx + c)^2 + y_0^2 - 2(mx + c)y_0 = r^2 \quad (5)$$

Solving (5) we have

$$x^2(1 + m^2) + x(2(mc - x_0 - y_0)) + (x_o^2 + y_0^2 + c^2 - r^2 - 2y_0c) = 0 \quad (6)$$

We can solve (6) using quadratic equation as shown below

$$\frac{-b \pm \sqrt{b^2 - 4ac}}{2a} \quad (7)$$

Where

$$a = (1 + m^2)$$
$$b = 2(mc - x_0 - y_0)$$
$$c = x_o^2 + y_0^2 + c^2 - r^2 - 2y_0c$$

This POI calculation is done for each threat DA pair. After calculating POIs we calculate time to DA using (8)

$$\text{Time} = \frac{\text{Distance}}{\text{Speed}} \quad (8)$$

Where Distance is calculated using Euclidian (9)

$$\sqrt{\sum_{i=1}^{n}(POI_i - TP_i)^2} \quad (9)$$

Where TP= target Position and i ∈ {1, 2}.

The target $T_n$ that ahs lesser time to $DA_i$ and is headed towards that $DA_i$ will have higher intent value for $DA_i$ as compared to a threat $T_m$ that has lesser time to $DA_i$ at a time stamp $t_0$. Figure 5 shows block diagram of proposed model.

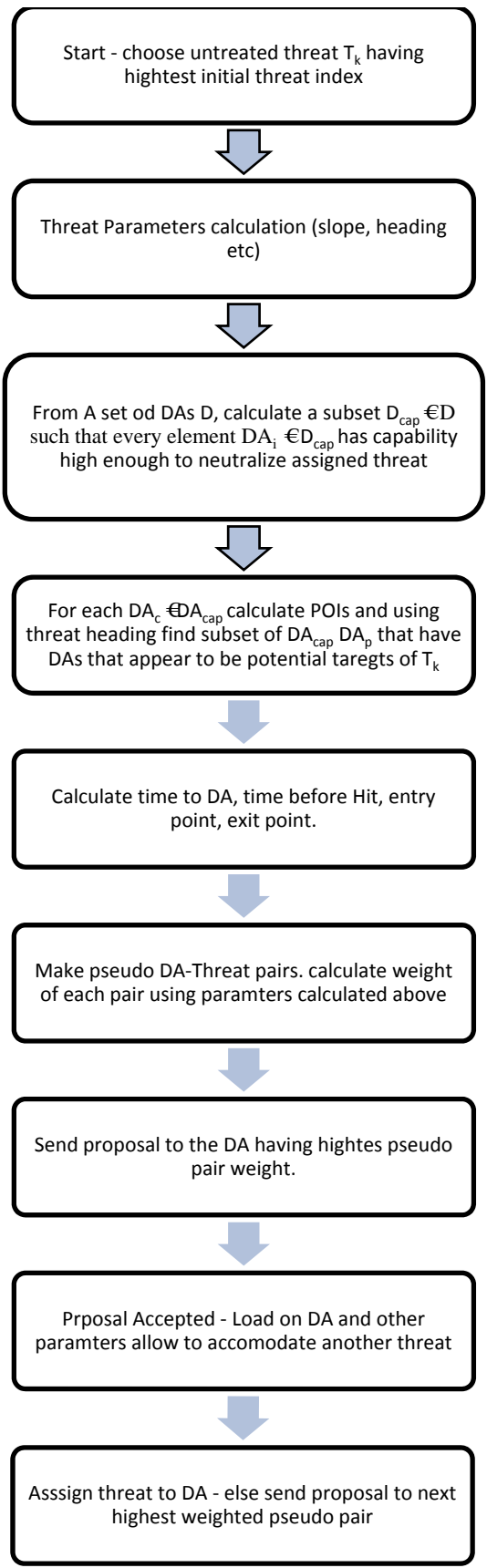

Figure 5: Proposed Solution

## C. Weapon Allocation Model

Once threat-DA pairing is done, next step is to assign members of a set of threats to the members of a set of Weapon Systems (WSs) present in assigned $DA_i$ i.e. creating matches between the elements of two disjoint sets.

Since threat and weapon libraries are correlated, using this correlation, this process creates a preference list of WS in general for the assigned threat showing the WSs that are good enough in terms of capability to neutralize assigned threat. Algorithm then searches for matching WSs from the set of WSs in hand. Using threat parameters like heading, course, direction etc, WA finds a subset of WSs found in previous set that have or are expected to have this threat in their range at some timestamp $t_0$. Let this set be represented by $WS_p$. For each WS' belonging to $WS_p$, a temporary pairing of threat and WS' is made. For each pair, algorithm calculates time to WS', distance from WS', tome of flight (TOF for WS'), required elevation angle for WS', lead calculations and launch points based on velocity of threat and TOF of WS'. For each temporary pair, algorithm calculates the weight of pair using a parametric weighted equation. A proposal is sent to the weapon system WS' of selected temporary pseudo pair. If it is accepted by the WS', threat is assigned to WS' else a new proposal is created and sent to the weapon system WS' belonging to next pseudo pair. Figure 6 shows the concept of entry point, exit points along with POI calculation for WS. The basic mathematics for WS – threat POI calculations is 50% same as that of DA-threat POI calculation. Since WSs have their own sweep and start angle. While calculation POIs, there is a need to consider WS parameters like Arc boundaries, sweep angle, start angle, elevation angle, field of fire etc.

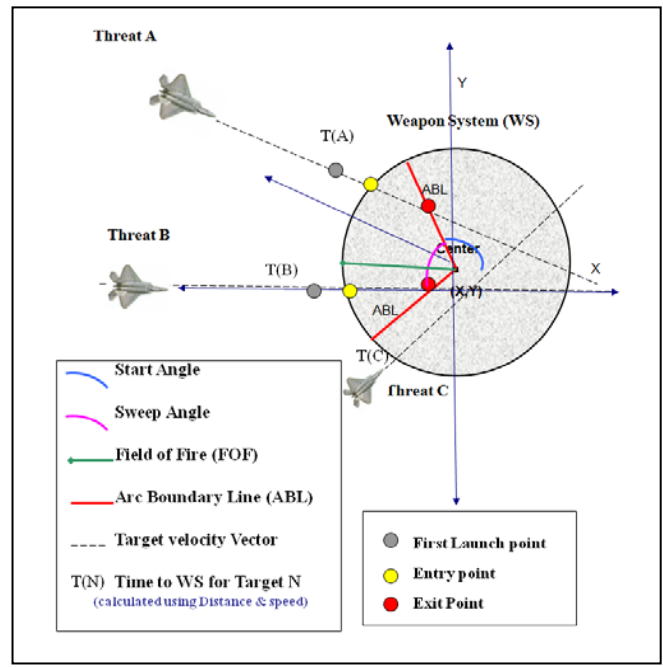

Once POIs are calculated, actual matching is done. This matching is subject to following constraints:

At most two threats can be schedules on a WS. i.e one threat can be locked and one can wait in the queue. For a WS j this can be expressed as:

$$\sum_{k=1}^{K} Scheduled_{j,k} \leq 2 \quad (10)$$

and

$$\sum_{j=1}^{J} Locked_{j,k} = 1 \quad (11)$$

Where $j \in WS_i$, $K \stackrel{\text{def}}{=}$ Number of Threats and. $WS_i$ shows the weapon set of $DA_i$ and

$$Scheduled_{j,k} = \begin{cases} 1 & if \ j \in WS_i \ and \ proposal_{i,k} = 1 \\ & and \ Porposal_{j,k} = 1 \\ 0 & Otherwise \end{cases}$$

and

$$Locked_{j,k} = \begin{cases} 1 & if \ Assign_{j,k} = 1 \ and \\ & k \ is \ not \ enqueued \ at \ j \\ 0 & Otherwise \end{cases}$$

The outcome each proposal is Boolean, it can either be 1 or 0 as given by equation 12.

$$proposal_{j,k} = \begin{cases} 1 & if \ j \in WS_i \ where \ WS_i \ shows \ WSs \ of \ DA \ i \\ & for \ which \ Porposal_{i,k} \ is \ accepted \ and \ Capability \\ & Index \ of \ WS_i > Required \ Minimum \ Value \\ 0 & Otherwise \ (no \ proposal \ is \ sent) \end{cases} \quad (12)$$

## IV. TESTING AND ANALYSIS

Different scenarios were created and tested on a system implementing this solution. Before running actual scenarios, we do the typical battlefield pre-processing simulation tasks that including DA definition, Weapon deployment and pairing with DA and communication server configuration. Communication server is used to make TEWA and simulator communicate through sockets.

After this battlefield pre-processing, we generate different scenarios to test the optimality of implemented system. These scenarios may range from relaxed K target scenario against J WSs to extremely stressful scenarios where time for scheduling is less or partial information is available to schedule targets. Scenarios can be designed to force starvation or over utilization probability increase explicitly. Cost analysis and damage calculations of this system show that although this algorithm is computation intensive but, even under stressful conditions, it succeeds in coming up with near optimal solution. Table 1 shows summary of few simulations. The main focus of this paper was on proposed model. We shall focus on our results and analysis in our next paper.

## V. CONCLUSION

A Novel Two-Staged Dynamic Decision Support based Optimal Threat Evaluation and Defensive Resource Scheduling Algorithm for Multi Air-borne threats is presented that correlates threat and weapon libraries to improve the solution quality. Opportunity, Proximity parameters along with DA characteristic and weapon/target parameters have been explored and used in a most prolific way. For optimality, these parameter weights are kept configurable. This paper explains the main optimization steps required to react to changing complex situations and provide near optimal weapon assignment for a range of scenarios. Proposed algorithm uses a variant of many-to-many Stable Marriage Algorithm (SMA) to solve Threat Evaluation (TE) and Weapon Assignment (WA) problem. TE corresponds to Threat Ranking and Threat-Asset pairing while WA corresponds to finding to best WS for each potential target using a flexible dynamic weapon scheduling algorithm, allowing multiple engagements using shoot-look-shoot strategy. Analysis part of this paper shows that this new approach to TEWA computes near-optimal solution for a range of scenarios.

TABLE I. SUMMARY OF FEW SIMULATIONS

| TE WA | *Number of Threats is less* | *Number of Threats is greater* | *Relaxed one-to-one case* |
|---|---|---|---|
| TE | K= 5 ; I= 10 ; All threats {T₁, T₂ … T₅₀}were allocated to best candidate DA within few milli-seconds. Some DAs were idle, some had balanced amount of load. | K= 50 ;I= 10 ; All threats {T1, T2 … T50}were allocated to best candidate DA within few seconds. For a smooth deployment, all DAs were engaged in TEWA. Few DAs were found idle when deployemnt is introduced with unbalanced WS assignment. | K= 10 ; I= 10 ; All threats {T1, T2 … T10}were allocated to best candidate DA within few milli-seconds. For a smooth deployment, all DAs were engaged in TEWA. Few DAs were found idle when deployemnt is introduced with unbalanced WS assignment. |
| WA | J= 10 ; K= 5 ; All threats {T₁, T₂ … T₅₀}were allocated to best candidate WS within few milli-seconds. The WS locked one target at a time. Few WS stayed idle. No resouce conflict found. System executed in subtractive mode, minimizing the expected survival time of each Threat. | J= 10 ; K= 50 ; All threats {T₁, T₂ … T₅₀}were allocated to best candidate WS within few seconds. The WS locked one target at a time and allowed to have one in its associated queue. Few targets stayed in a queue. System shifted to preferential mode, maximizing the total survival value of each DA. | J= 10 ; K= 50 ; All threats {T₁, T₂ … T₅₀}were allocated to best candidate WS within few seconds. The WS locked one target at a time and allowed to have one in its associated queue. System continued to stay in the same mode it was already executing in. |